\documentclass[10pt,twocolumn]{article}

\usepackage[margin=0.75in]{geometry}
\usepackage{times}
\usepackage{graphicx}
\usepackage{amsmath}
\usepackage{amssymb}
\usepackage{amsthm}
\usepackage{booktabs}
\usepackage{xcolor}
\usepackage{microtype}
\usepackage{hyperref}
\usepackage[capitalise,noabbrev]{cleveref}

\newtheorem{definition}{Definition}
\newtheorem{lemma}{Lemma}

\title{When Certificates Fail: A Unified Safety Framework for Embedded Neural Interface Models}

\author{
Jasmeet Singh Bindra\\
IKSMHA Center, IIT Mandi\\
\texttt{D24059@students.iitmandi.ac.in}
}

\begin{document}

\maketitle

\begin{abstract}
Formal robustness certificates for embedded neural-interface models can pass while task accuracy collapses: at perturbation budget $\epsilon{=}0.25$, EEGNet classification accuracy drops by 25.7\% under projected-gradient attack while the Lipschitz-style certificate remains valid for all 9 tested subjects. We argue that this gap between mathematical certification and operational safety is one instance of a broader alignment failure in neural interfaces, where training objectives diverge from user welfare. We propose a unified empirical audit framework organized around three such failures: \emph{verification insufficiency}, in which certificates pass while task behavior degrades; \emph{proxy-fidelity divergence}, in which task-optimized representations damage neural signal structure (a time-domain auxiliary objective reduces reconstruction MSE by $0.1132$ while worsening spectral log-MSE); and \emph{latent information exfiltration}, in which public-task embeddings retain private attributes (subject identity recoverable at 48.1\% versus 6.7\% chance). We instantiate the framework on BCI Competition IV 2a and SEED-IV using multiple deep and classical EEG decoders, official session-level validation, null controls, and paired statistical tests. The verification gap persists across EEGNet, CSP+LDA, and FBCSP+LDA, and is therefore architecture-independent. Our results establish that operational safety auditing, not certificate verification alone, is necessary for responsible neural-interface deployment.
\end{abstract}

\section{Introduction}

Neural interfaces are moving from laboratory prototypes into clinical deployment across rehabilitation, assistive communication, neuroprosthetic control, and affective computing \cite{wolpaw2002bci,lebedev2006bmi,hochberg2012reach}. In these systems, embedded machine-learning models transform noisy neural measurements into safety-relevant decisions about intended movement, cognitive state, or device control \cite{lawhern2018eegnet,schirrmeister2017deep}. Errors in these decisions can affect user autonomy, communication, therapy, and privacy, even when the model is small enough to run on an edge device \cite{lopezbernal2021security,ienca2017rights}.

This makes neural-interface deployment an alignment problem in a concrete technical sense. The model is optimized against a tractable training objective: classification accuracy, reconstruction loss, or calibration performance, while the user's true welfare includes goals that are broader and harder to formalize: reliable intent expression, preservation of clinically meaningful signal structure, graceful degradation under perturbation, and protection of private neural attributes \cite{amodei2016concrete,ngo2022alignment}. A neural decoder can therefore succeed under its training proxy while failing the user-level objective that motivated the interface \cite{leike2017gridworlds}. This paper treats that mismatch as an engineering object, asking whether empirical safety audits can expose places where the learned proxy diverges from operational user welfare \cite{ngo2022alignment,ji2023alignment}.

Current safety work for neural interfaces is fragmented. Robustness studies test decoder stability under noise and adversarial manipulation \cite{liu2019universal,chen2022adversarial_artifact_detection}. Signal-fidelity studies test whether learned representations preserve physiologically meaningful structure \cite{lawhern2018eegnet,schirrmeister2017deep}. Privacy studies test whether embeddings reveal sensitive attributes beyond the intended task \cite{ienca2017rights,lopezbernal2021security}. Each line of work is necessary, but treating them independently obscures a common failure pattern: the deployed model satisfies one narrow specification while violating a broader safety requirement \cite{amodei2016concrete,hendrycks2021unsolved}. A robustness certificate can validate a mathematical inequality while saying nothing about task failure; a classifier can optimize motor-imagery accuracy while distorting neural information needed for faithful communication; and an encoder can support an intended task while retaining private subject identity in its latent representation. No existing framework connects these three phenomena: robustness certification, proxy-fidelity divergence, and latent privacy leakage, as related manifestations of neural-interface alignment failure.

The sharpest instance of this gap arises in formal verification and robustness certification \cite{hein2017formal,weng2018clever,fazlyab2019quadratic}. Lipschitz-style bounds, spectral constraints, margin certificates, and related verification tools provide sufficient mathematical statements about how much a model output can change under bounded input perturbations \cite{cisse2017parseval,weng2018clever,fazlyab2019lipsdp}. Such certificates are attractive for safety-critical systems because they convert an otherwise empirical robustness question into a checkable property of the model class or learned parameters \cite{fazlyab2019quadratic}. However, a certificate can be true and still operationally uninformative if it is too conservative, if it certifies output movement rather than task correctness, or if the certified property is not aligned with the deployment hazard \cite{hein2017formal,weng2018clever}. We empirically demonstrate this failure mode in embedded EEG decoding: conservative output-sensitivity checks pass across tested subjects and perturbation budgets, while adversarially perturbed inputs produce sharp drops in motor-imagery classification accuracy. In the strongest version of the result, official-session EEGNet and classical CSP/FBCSP decoders lose substantial accuracy under bounded projected-gradient attacks even as the conservative certificate-style pass condition remains satisfied. A mathematically valid certificate can fail to flag the operational hazard that matters to a user or regulator.

We propose a unified empirical audit framework grounded in three AI safety failure modes and instantiate it on public EEG/BCI data \cite{amodei2016concrete,leike2017gridworlds,hendrycks2021unsolved}. First, the verification audit (E1) targets an operational analogue of deceptive alignment: a model or certificate appears safe under the audited formal property while behavior degrades under task-relevant stress \cite{hubinger2019risks}. Second, the proxy-fidelity audit (E2) targets specification gaming: a representation can optimize the proxy task while failing to preserve neural signal properties required by a richer fidelity objective \cite{amodei2016concrete,leike2017gridworlds}. Third, the privacy leakage audit (E3) targets latent information exfiltration: embeddings trained for a public task may retain recoverable private attributes, such as subject identity, even when those attributes are not part of the intended interface output \cite{ienca2017rights,lopezbernal2021security}. We validate these audits across Braindecode EEGNet, CSP+LDA, and FBCSP+LDA on official BCI Competition IV 2a train-to-evaluation protocols, SEED-IV session-holdout privacy probes with four probe families, label and private-attribute permutation controls, physiological perturbation stress tests, and paired statistical analyses \cite{lawhern2018eegnet,ang2008filterbank,jayaram2018moabb,schirrmeister2017deep}. Across this evidence base, the anchoring result is that conservative certificates can pass while task behavior collapses, which motivates auditing verification, fidelity, and privacy as linked safety requirements rather than isolated benchmarks.

\paragraph{Contributions.}
(1)~We formulate embedded neural-interface safety as an alignment problem and define three failure modes: verification insufficiency, proxy-fidelity divergence, and latent information exfiltration, grounded in a common taxonomy derived from the AI safety literature.
(2)~We instantiate this taxonomy as an empirical audit on public EEG/BCI datasets, demonstrating that conservative Lipschitz-style robustness certificates pass while classification accuracy collapses by up to 25.7\% under bounded adversarial perturbation; a result that holds across EEGNet, CSP+LDA, and FBCSP+LDA decoders.
(3)~We show that proxy-fidelity divergence is objective-dependent (time-domain and spectral fidelity cannot be simultaneously preserved) and that public-task embeddings leak private subject identity at 48.1\% versus 6.7\% chance, establishing that operational auditing across all three failure modes is necessary for responsible deployment.

\Cref{sec:related} reviews related work. \Cref{sec:threat_model} defines the threat model and formalizes three failure modes. \Cref{sec:methodology} presents the audit methodology and null controls. \Cref{sec:experiments} describes datasets, architectures, and experimental results. \Cref{sec:discussion} discusses governance implications, and \Cref{sec:conclusion} concludes with open directions.

\section{Related Work}\label{sec:related}

\subsection{AI Safety and Alignment Theory}

This work is motivated by a technical-to-governance bridge: safety research should be technically grounded while remaining legible to real-world deployment practice and policy. We organize the audit around three safety-relevant axes. \emph{Verification} asks whether system behavior can be bounded, audited, or certified before deployment. \emph{Frontier safety} asks whether increasingly capable learned systems can fail in ways not captured by ordinary benchmark performance. \emph{Agent governance} asks how technical evidence should inform oversight when deployed models act on behalf of users or device operators. Neural interfaces are not frontier foundation models, but they instantiate the same structural problem at a smaller and more embodied scale: a learned decoder mediates between human intent and machine action.

The alignment literature provides the conceptual vocabulary for this problem. Concrete accident risks such as reward hacking and distribution shift arise whenever a system optimizes an imperfect real-world objective \cite{amodei2016concrete}. Inner misalignment can cause a learned model to optimize a mesa-objective different from the training objective \cite{hubinger2019risks}, and goal misgeneralization can produce deceptive or power-seeking behavior as systems scale \cite{ngo2022alignment}. We apply these concepts conservatively: EEG decoders are not strategic agents, but they can still optimize proxies, pass narrow certification tests, and fail user-level safety objectives.

This paper therefore translates alignment concepts into audit requirements for embedded neural-interface models. Specification gaming appears when proxy task performance does not imply neural fidelity. Deceptive alignment is treated as an operational analogy: a certificate can indicate safety under one formal property while the deployed behavior degrades under task-relevant perturbations. Loss of control and corrigibility motivate asking whether users, clinicians, or regulators can detect and correct such failures before the interface acts on them \cite{soares2015corrigibility,ngo2022alignment}. The contribution is not a new general theory of alignment, but a domain-specific audit methodology that makes these failure modes measurable in neural data. These three axes map directly onto the audits that follow: verification motivates E1 (certificate gap), frontier safety motivates E2 (proxy-fidelity divergence), and agent governance motivates E3 (latent privacy leakage).

\subsection{Neural Interface Security}

Security research on brain-computer interfaces has primarily emphasized adversarial access, signal manipulation, privacy compromise, and attacks across the BCI life cycle. Lopez Bernal et al. survey BCI security threats and organize attacks around acquisition, processing, transmission, classification, and feedback stages \cite{lopezbernal2021security}. This line of work is essential because neural interfaces expose unusually sensitive channels: attackers may target not only software and network surfaces, but also neural signals, stimulation loops, and inferred mental or health states. More recent EEG adversarial-ML studies show that BCI classifiers can be attacked by universal or detectable adversarial perturbations \cite{liu2019universal,chen2022adversarial_artifact_detection}.

Governance and neuroethics work broadens the concern from malicious attackers to institutional responsibility, user autonomy, consent, and the regulation of neural data \cite{ienca2017rights,ienca2016hacking,sirbu2025regulating_bci}.

Our safety framing differs from the standard cybersecurity framing. Cybersecurity usually begins with an external adversary: someone steals data, injects noise, manipulates a signal, or compromises a device. We include adversarial perturbations and privacy probes, but our main question is broader: can the model's learned objective become misaligned with user welfare even without an external attacker? A motor-imagery classifier may be fragile under bounded perturbations; a representation may preserve enough information to identify the user; and a reconstruction objective may improve one fidelity metric while damaging another. These are safety failures even when no network intrusion occurs. The audit framework therefore complements BCI security surveys by treating robustness, fidelity, and privacy as coupled alignment risks.

\subsection{Verification and Robustness Certification}

Adversarial robustness and certification have a mature literature in computer vision and general machine learning. Szegedy et al. showed that neural networks can be vulnerable to small adversarial perturbations \cite{szegedy2013intriguing}, and Madry et al. formalized projected-gradient adversarial training through a robust-optimization lens \cite{madry2017towards}. Certification methods attempt to move beyond attack-specific evaluation by proving that model outputs or decisions cannot change within a specified perturbation set. Lipschitz-based methods are especially relevant here because they provide sufficient bounds on output sensitivity: if the network's Lipschitz constant and the perturbation radius are bounded, then output movement is bounded. Hein and Andriushchenko gave formal classifier robustness guarantees using cross-Lipschitz reasoning \cite{hein2017formal}; Weng et al. proposed CLEVER as a scalable local Lipschitz robustness estimate \cite{weng2018clever}; and Fazlyab et al. developed semidefinite and quadratic-constraint methods for robustness and Lipschitz analysis \cite{fazlyab2019quadratic,fazlyab2019lipsdp}.

These tools are powerful, but most operational validation has been concentrated in image classification, standard adversarial benchmarks, or abstract neural-network verification settings. Neural interfaces add constraints that are easy to miss in generic robustness work: inputs are physiological time series, perturbations may correspond to sensor noise or channel failure, task labels are noisy proxies for intent, and the harms include autonomy, communication failure, and leakage of private neural attributes. A certificate that bounds logit movement may be valid yet too loose to certify task correctness, and a perturbation that is small in norm may still be consequential for a user-facing decoder.

This is the gap E1 targets. Rather than rejecting verification, we test its operational meaning: we compare conservative certificate-style output-sensitivity checks against PGD task failure, classical BCI baselines, and physiological stress tests. The result is direct: a certificate can pass while the task fails, so verification should be reported alongside adversarial and statistical audits rather than as a standalone safety claim.

\section{Threat Model and Problem Formulation}\label{sec:threat_model}

\subsection{Neural Interface as an Alignment Problem}

We model an embedded neural interface as a stochastic pipeline from neural measurements to an action or decision. Let $x \in \mathcal{X}$ denote a preprocessed neural signal segment, such as an EEG epoch. An encoder $f_{\theta}: \mathcal{X} \rightarrow \mathcal{Z}$ maps the signal into a latent representation
\begin{equation}
    z = f_{\theta}(x),
\end{equation}
and a downstream classifier, decoder, or controller $g_{\phi}: \mathcal{Z} \rightarrow \mathcal{A}$ maps this representation to an output
\begin{equation}
    a = g_{\phi}(z) = g_{\phi}(f_{\theta}(x)).
\end{equation}
The output $a$ may be a motor-imagery class, communication symbol, affective-state estimate, stimulation decision, or device-control command. During training, the model is optimized with respect to a tractable objective
\begin{equation}
    \mathcal{L}_{\mathrm{train}}(\theta,\phi)
    =
    \mathbb{E}_{(x,y)\sim \mathcal{D}_{\mathrm{train}}}
    \left[
    \ell(g_{\phi}(f_{\theta}(x)), y)
    \right],
\end{equation}
possibly augmented with regularization, reconstruction, or privacy terms. The user's true welfare, however, is not identical to this loss. We write it abstractly as $W(x,a)$: a deployment-level utility that depends on whether the output respects the user's intent, preserves clinically or communicatively meaningful neural information, degrades safely under perturbation, and avoids exposing sensitive attributes. The central alignment problem is therefore that minimizing $\mathcal{L}_{\mathrm{train}}$ need not maximize $W(x,a)$.

\begin{definition}[Verification insufficiency]
Let $C(f_{\theta},\epsilon)\in\{\mathrm{true},\mathrm{false}\}$ be a robustness certificate for perturbations $\delta$ satisfying $\|\delta\|\leq \epsilon$, and let $S(f_{\theta},g_{\phi},\epsilon)$ denote the corresponding task-level safety property, such as preservation of the correct class under the same perturbation budget. Verification insufficiency occurs when
\begin{equation}
    C(f_{\theta},\epsilon)=\mathrm{true}
    \quad \mathrm{and} \quad
    S(f_{\theta},g_{\phi},\epsilon)=\mathrm{false}.
\end{equation}
\end{definition}

\begin{lemma}[Lipschitz output-sensitivity bound]
Suppose $m_{\theta,\phi}=g_{\phi}\circ f_{\theta}$ is $L$-Lipschitz with respect to a norm $\|\cdot\|$, so that
\begin{equation}
    \|m_{\theta,\phi}(x_1)-m_{\theta,\phi}(x_2)\|
    \leq L\|x_1-x_2\|
\end{equation}
for all $x_1,x_2\in\mathcal{X}$. Then for any perturbation $\delta$ with $\|\delta\|\leq \epsilon$,
\begin{equation}
    \|m_{\theta,\phi}(x+\delta)-m_{\theta,\phi}(x)\|
    \leq L\epsilon.
\end{equation}
\end{lemma}

The lemma is valid but narrow: it bounds output movement, not task correctness. A margin-based classifier guarantee requires the output bound to be small relative to the decision margin; when the Lipschitz constant is loose, as it typically is for product-of-layer estimates, the certificate $C$ holds vacuously while the safety property $S$ fails. This gap is not a mathematical error; it is a structural limitation of sufficient-condition certificates applied to deployment-level safety claims.

\begin{definition}[Proxy-fidelity divergence]
Let $\mathcal{L}_{\mathrm{proxy}}$ be a training loss for a downstream task, and let $\widehat{x}=r_{\psi}(z)$ be a reconstruction or decoded signal from the learned representation. Let $F(x,\widehat{x})$ be a fidelity metric, where larger values indicate better preservation of task-relevant neural structure. Proxy-fidelity divergence occurs when optimizing $\mathcal{L}_{\mathrm{proxy}}$ improves the proxy objective but degrades or fails to improve $F(x,\widehat{x})$ relative to a representation trained with an explicit fidelity objective.
\end{definition}

\begin{definition}[Latent information exfiltration]
Let $p\in\mathcal{P}$ denote a private attribute, such as subject identity, that is not part of the intended interface output. Let $h:\mathcal{Z}\rightarrow\mathcal{P}$ be an attribute probe trained on latent representations. Latent information exfiltration occurs when
\begin{equation}
    \mathrm{Acc}(h(z),p) \gg \frac{1}{|\mathcal{P}|},
\end{equation}
under a held-out evaluation protocol, indicating that private information remains recoverable from $z$ above chance.
\end{definition}

These definitions separate three different ways in which $\mathcal{L}_{\mathrm{train}}$ can fail to capture $W(x,a)$. E1 asks whether formal sensitivity checks imply task safety. E2 asks whether task optimization preserves neural fidelity. E3 asks whether task embeddings avoid private-attribute leakage.

\subsection{Failure Mode Taxonomy}

The three audits form a compact taxonomy of neural-interface alignment failures. E1, the verification gap, corresponds to an operational analogue of deceptive alignment and to a Goodhart failure on certificates. The model or verification artifact appears acceptable under the measured formal criterion, but the behavior that matters for the user deteriorates under perturbation. In our notation, the certificate $C(f_{\theta},\epsilon)$ remains true, yet the task-level safety property $S(f_{\theta},g_{\phi},\epsilon)$ is false. The empirical prediction is that output-sensitivity bounds may pass across perturbation budgets while adversarial or physiological stress tests reveal large drops in classification accuracy, prediction stability, or clean-correct retention.

E2, proxy-fidelity divergence, corresponds to specification gaming or reward misspecification. The training objective rewards a proxy, such as motor-imagery classification accuracy, because it is observable and convenient. But the user's welfare may depend on preserving richer neural information than the class label. The empirical prediction is that a representation optimized only for $\mathcal{L}_{\mathrm{proxy}}$ can achieve plausible task performance while performing poorly under separately measured fidelity metrics. If the taxonomy is correct, adding an explicit fidelity term should shift the representation in metric-dependent ways: it may improve time-domain reconstruction, spectral preservation, or correlation, but not necessarily all at once and not necessarily without a task-accuracy tradeoff.

E3, latent privacy leakage, corresponds to latent information exfiltration and unintended capability acquisition. The encoder is trained for a public task, but its latent state may contain information that supports an unintended private task. This is not necessarily malicious behavior by the model; it is an emergent capability of the representation. The empirical prediction is that private probes $h$ should recover attributes such as subject identity from $z$ above the chance baseline $1/|\mathcal{P}|$, and that sanitization mechanisms should reduce this leakage while preserving as much public-task utility as possible. A private-attribute permutation control should collapse probe performance to chance, distinguishing real leakage from probe artifacts.

Together, the taxonomy yields a testable prediction: \emph{a serious safety audit must be multi-objective}. A single clean accuracy number cannot distinguish robust task behavior from brittle task behavior, faithful representations from proxy-optimized representations, or public-task embeddings from privacy-leaking embeddings. The experiments that follow test this prediction by applying E1, E2, and E3 across deep and classical EEG decoders, official evaluation protocols, null controls, and paired statistical analyses.

\section{Methodology}\label{sec:methodology}

\begin{figure*}[t]
\centering
\includegraphics[width=0.92\textwidth]{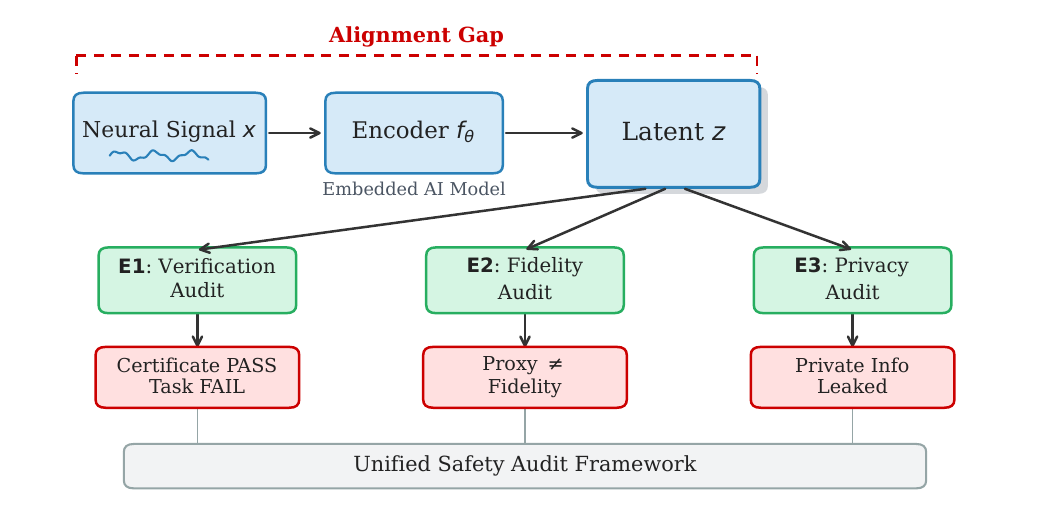}
\caption{Unified safety audit framework. Embedded neural-interface models map neural signals through a learned encoder into latent states, where alignment gaps can surface as verification insufficiency, proxy-fidelity divergence, or latent privacy leakage.}
\label{fig:framework_overview}
\end{figure*}

\subsection{Verification Audit (E1)}

The verification audit evaluates whether a conservative output-sensitivity certificate is operationally meaningful for EEG decoders. For neural models, we compute an upper bound on the model Lipschitz constant by multiplying layerwise operator bounds. In a compact spectral-normalized classifier, this uses the spectral norms of linear and convolutional components together with conservative BatchNorm scale factors. The resulting constant $L$ gives the bare lemma check in \Cref{sec:threat_model}: for any perturbation $\|\delta\|\leq\epsilon$, observed output movement should satisfy
\begin{equation}
    \|m_{\theta,\phi}(x+\delta)-m_{\theta,\phi}(x)\| \leq L\epsilon .
\end{equation}
This check is intentionally treated as a sufficient output-movement audit, not as a task-safety guarantee.

We test this certificate with a multi-epsilon sensitivity audit. The all-subject gradient audit used BCI Competition IV 2a subjects A01T through A09T, epochs $=8$, epsilons $\{0.05,0.1,0.25,0.5\}$, and $16$ audit trials. For each model, we evaluate random $L_2$ perturbations and gradient-aligned perturbations. Random perturbations estimate typical local sensitivity under isotropic directions, while gradient-aligned perturbations approximate sharper local directions by increasing output movement. This separates ordinary perturbation response from a more adversarial output-sensitivity probe.

We then add projected-gradient descent (PGD) classification attacks to test task-level failure. The PGD attack grid uses epsilons $\{0,0.05,0.1,0.25,0.5,1.0\}$, $40$ PGD steps, and $2$ restarts. This grid is applied to the tiny spectral classifier, Braindecode EEGNet, and classical CSP+LDA/FBCSP+LDA baselines. In the official-session BCI IV 2a validation, EEGNet is trained for $80$ epochs on A01T through A09T and evaluated on the corresponding A01E through A09E sessions using MOABB labels. Classical CSP+LDA uses the $8$-$30$ Hz band with $4$ filters per class and $16$ features; FBCSP+LDA uses bands $4$-$8$, $8$-$12$, $12$-$16$, $16$-$20$, $20$-$24$, $24$-$28$, $28$-$32$, and $32$-$38$ Hz, with $4$ filters per class and $128$ features.

For each perturbation budget, we report both the bare lemma pass/fail result and a margin-based certified-safe fraction. Let $\gamma_i$ be the clean logit margin for a correctly classified example $i$, defined as the true-class logit minus the largest competing logit. Under the conservative logit-movement bound used in the experiments, a clean-correct example is counted as certified safe only when its margin exceeds approximately $\sqrt{2}L\epsilon$. The certified-safe fraction is the fraction of examples satisfying this margin condition. This differs from the bare lemma check: the lemma only verifies that outputs do not move more than $L\epsilon$, while task safety requires the output movement to be small enough that the decision cannot cross a class boundary.

\subsection{Proxy-Fidelity Audit (E2)}

The proxy-fidelity audit tests whether a representation optimized for a public task preserves neural signal information under a separately defined fidelity metric. For each BCI IV 2a subject, we compare two models with the same latent-bottleneck design: a classification-only encoder trained for motor-imagery classification, and a joint encoder trained with classification plus reconstruction. The core contrast is not between different architectures, but between objectives under a matched representation constraint.

The classification-only condition is evaluated with a frozen reconstruction probe. After training the classifier, the encoder is frozen and a decoder is trained from its latent representation to reconstruct the input epoch. This measures how much recoverable signal information remains in a task-trained latent space without allowing the encoder to adapt to the reconstruction objective. The joint condition trains the encoder and decoder together, so reconstruction pressure can shape the latent representation directly. In the official-session EEGNet E2 runs, subjects A01T through A09T are trained on the official training sessions and evaluated on A01E through A09E; the runs use $40$ classifier epochs and $40$ probe epochs.

We evaluate three fidelity metrics. Time-domain mean squared error measures pointwise reconstruction error between $x$ and $\widehat{x}$. Reconstruction correlation measures linear agreement between reconstructed and original signal structure. Spectral log-MSE measures discrepancy in log spectral power and is used because EEG decoding can depend strongly on frequency-domain structure. These metrics are intentionally reported separately, since improvement in one fidelity metric need not imply improvement in another.

We test three joint objectives. The time-domain-only objective uses cross-entropy plus a reconstruction MSE term with weight $1.0$. The full-spectrum objective uses
\begin{equation}
    \mathcal{L}
    =
    \mathcal{L}_{\mathrm{CE}}
    + 1.0\,\mathcal{L}_{\mathrm{time\_mse}}
    + 0.1\,\mathcal{L}_{\mathrm{full\_spectral}},
\end{equation}
and the bandpower objective uses the same $1.0$ time-MSE weight and $0.1$ spectral weight, but computes spectral loss over theta, alpha/mu, low-beta, high-beta, beta, and low-gamma bands. This design lets E2 ask whether explicit fidelity terms repair the proxy-fidelity gap, and also whether the repair is metric-specific.

\subsection{Privacy Leakage Audit (E3)}

The privacy audit tests whether embeddings trained for a public neural task retain private subject information. The public task is SEED-IV emotion recognition using the \texttt{de\_LDS} feature representation. Each sample is a $310$-dimensional feature vector corresponding to $62$ EEG channels and $5$ frequency bands. The private attribute is closed-set subject identity over $15$ subjects. The main split is a session-holdout protocol: sessions $1$ and $2$ are used for training, and session $3$ is held out for evaluation. This yields $37{,}575$ windows, with $25{,}245$ training windows and $12{,}330$ test windows. The split is used because random window-level splits can overstate generalization by mixing nearby trials or sessions across train and test.

We first train a task encoder for the public emotion label, then extract latent embeddings $z=f_{\theta}(x)$ for train and test windows. Privacy probes are then fit to predict subject identity from $z$. The stronger E3 suite uses four probe families: linear logistic regression, $k$-nearest neighbors, random forest, and multilayer perceptron. In the full stronger-probe runs, seeds are $\{7,13,21\}$ and the task encoder is trained for $20$ epochs. Linear probes use the full training set with $n_{\mathrm{train\_fit}}=25{,}245$. The kNN, random-forest, and MLP probes use class-balanced training subsets with $n_{\mathrm{train\_fit}}=6{,}000$ for tractability. The random forest uses $80$ trees, and the MLP uses a maximum of $40$ iterations. Probe seeds are held constant across privacy states within a run so that changes in measured leakage are attributable to the sanitizer rather than stochastic probe refits.

We evaluate two simple sanitization mechanisms. Projection removal estimates subject-discriminative directions in the latent space and removes $0$, $1$, $2$, $4$, $8$, or $12$ such directions. Gaussian latent noise adds noise at scales $0$, $0.1$, $0.25$, $0.5$, and $1.0$ after embedding standardization. For each sanitized state, we report public emotion accuracy, private subject-probe accuracy, and leakage reduction relative to the unsanitized baseline. The audit succeeds as a privacy diagnostic if subject identity is recoverable above the $1/15$ chance level and if real-label leakage behaves differently from private-attribute permutation controls.

\subsection{Null Controls and Statistical Framework}

Null controls are used to distinguish real signal from loader, split, or probe artifacts. For E1, the label-permutation control permutes motor-imagery training labels after the train/test split while leaving test labels true. The control uses A01T through A09T, epsilons $\{0,0.05,0.1,0.25,0.5,1.0\}$, $40$ PGD steps, and $2$ restarts. For E2, the label-permutation control also permutes classification labels for classifier and joint training, while reconstruction targets remain true neural epochs. This tests whether reconstruction improvements are merely a consequence of class-label semantics. For E3, the private-attribute permutation control permutes subject labels within the real session-holdout split while keeping true emotion labels, so a valid privacy signal should collapse to chance when the private labels are randomized.

Statistical summaries are generated from the canonical CSV outputs rather than hand-copied from logs. For each paired effect, we report a percentile bootstrap $95\%$ confidence interval for the mean effect and a paired sign-flip permutation test against zero. Directional $p$-values follow the expected sign of the effect, while two-sided $p$-values are retained in the machine-readable CSV. Benjamini-Hochberg false-discovery-rate correction is applied within each analysis family.

The sign-flip test has discrete resolution at the available sample sizes. For BCI IV 2a subject-paired analyses with $n=9$, the smallest exact one-sided $p$-value is $0.001953$ and the smallest exact two-sided $p$-value is $0.003906$. For the stronger E3 multiseed analyses with $n=3$, the smallest exact one-sided $p$-value is $0.125$ and the smallest exact two-sided $p$-value is $0.25$. E3 inferential claims are therefore treated as descriptive multiseed evidence rather than conventionally significant hypothesis tests.

\section{Experiments and Results}\label{sec:experiments}

\subsection{Datasets and Architectures}

We evaluate the audit framework on two public EEG datasets. The BCI Competition IV 2a experiments use the official four-class motor-imagery task with nine subjects, A01 through A09. The canonical BCI results use the official session-level protocol: each subject's training session is used for fitting the model, and the corresponding evaluation session is used for testing. Evaluation labels are loaded through MOABB's BNCI2014\_001 interface, because the locally converted evaluation-session event files retain unknown cue codes rather than true class labels. This official-session protocol is used for the primary EEGNet PGD, CSP/FBCSP PGD, EEGNet proxy-fidelity, and physiological robustness experiments.

The privacy experiments use SEED IV. The public task is four-class emotion recognition, and the private attribute is closed-set subject identity over 15 subjects. The canonical E3 experiments use the \texttt{de\_LDS} feature representation, where each window is a 310-dimensional vector constructed from 62 EEG channels and 5 frequency bands. The split is session-holdout: sessions 1 and 2 are used for training, and session 3 is held out for evaluation. This split is stricter than random window-level splitting because it tests whether private information persists across recording sessions rather than merely across nearby windows.

We include both deep and classical EEG architectures. The primary deep decoder is Braindecode EEGNet with 2,484 trainable parameters in the E1 audit. The initial certificate machinery is also tested on a compact spectral-normalized EEG classifier. Classical baselines include CSP+LDA and FBCSP+LDA, providing EEG-literature controls that do not rely on deep representation learning. This mix lets the audit distinguish failures that are specific to one neural architecture from failures that persist across established decoding paradigms.

\subsection{E1: Robustness Verification Gap}

\begin{figure}[t]
\centering
\includegraphics[width=\columnwidth]{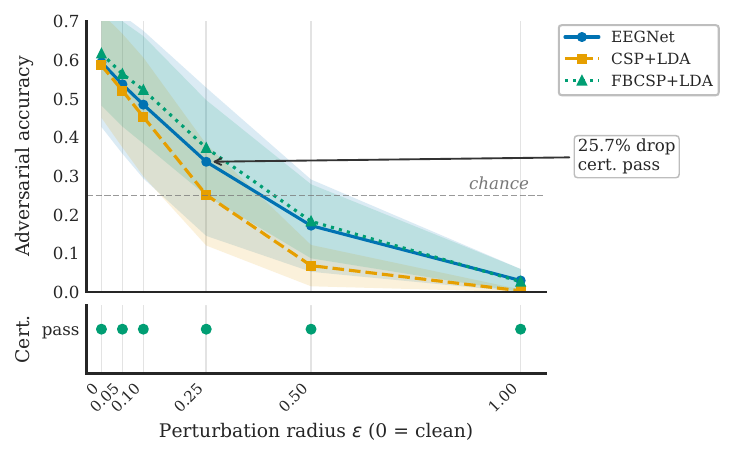}
\caption{Adversarial task accuracy collapses under bounded PGD perturbations while the EEGNet Lipschitz-style certificate check remains passed at every tested $\epsilon$. Shaded bands show $\pm 1$ SD across subjects.}
\label{fig:pgd_accuracy_drop}
\end{figure}

Across all nine official-session BCI IV 2a subjects, the conservative output-sensitivity check passes at every tested perturbation budget, but adversarial classification accuracy degrades sharply as $\epsilon$ increases. The margin-certified safe fraction is zero for every nonzero budget, confirming that the bare Lipschitz-style lemma is much weaker than a task-level safety certificate. Mean clean official-session accuracies are $0.5941 \pm 0.1673$ (EEGNet), $0.5860 \pm 0.1354$ (CSP+LDA), and $0.6169 \pm 0.1347$ (FBCSP+LDA), all well above the four-class chance level of $0.25$. Table~\ref{tab:pgd_accuracy_drops} reports the paired accuracy drops, and Figure~\ref{fig:pgd_accuracy_drop} shows the accuracy-versus-$\epsilon$ curve.

\begin{table}[t]
\centering
\caption{PGD adversarial accuracy drops under the official BCI IV 2a session protocol.}
\label{tab:pgd_accuracy_drops}
\footnotesize
\begin{tabular}{llcc}
\toprule
Model & $\varepsilon$ & Accuracy drop & $n$ \\
\midrule
EEGNet & 0 & $0.0000 \pm 0.0000$ & 9 \\
EEGNet & 0.05 & $0.0571 \pm 0.0230$ & 9 \\
EEGNet & 0.10 & $0.1096 \pm 0.0407$ & 9 \\
EEGNet & 0.25 & $\mathbf{0.2573 \pm 0.0583}$ & 9 \\
EEGNet & 0.50 & $0.4225 \pm 0.0833$ & 9 \\
EEGNet & 1.00 & $\mathbf{0.5644 \pm 0.1507}$ & 9 \\
\midrule
CSP+LDA & 0 & $0.0000 \pm 0.0000$ & 9 \\
CSP+LDA & 0.05 & $0.0652 \pm 0.0162$ & 9 \\
CSP+LDA & 0.10 & $0.1331 \pm 0.0244$ & 9 \\
CSP+LDA & 0.25 & $\mathbf{0.3349 \pm 0.0541}$ & 9 \\
CSP+LDA & 0.50 & $0.5177 \pm 0.0980$ & 9 \\
CSP+LDA & 1.00 & $\mathbf{0.5829 \pm 0.1328}$ & 9 \\
\midrule
FBCSP+LDA & 0 & $0.0000 \pm 0.0000$ & 9 \\
FBCSP+LDA & 0.05 & $0.0529 \pm 0.0216$ & 9 \\
FBCSP+LDA & 0.10 & $0.0941 \pm 0.0296$ & 9 \\
FBCSP+LDA & 0.25 & $\mathbf{0.2442 \pm 0.0509}$ & 9 \\
FBCSP+LDA & 0.50 & $0.4340 \pm 0.0922$ & 9 \\
FBCSP+LDA & 1.00 & $\mathbf{0.5903 \pm 0.1250}$ & 9 \\
\bottomrule
\end{tabular}
\vspace{0.25em}
\parbox{\columnwidth}{\footnotesize \emph{Note.} Values are mean $\pm$ SD across subjects; bold marks the paper's moderate attack budget ($\varepsilon=0.25$) and the largest tested-budget drops.}
\end{table}

This finding is a gap, not a contradiction. The certificate checks whether observed output movement remains below $L\epsilon$; it does not check whether the true-class logit margin remains positive. The official-session EEGNet results therefore demonstrate that a decoder can satisfy the certified output-movement property while failing the user-relevant classification property, converting a formal safety claim into an operational audit question.

The classical baselines test whether this failure is peculiar to EEGNet. CSP+LDA and FBCSP+LDA are standard EEG decoding pipelines, and FBCSP is often a strong motor-imagery baseline. Both classical pipelines show substantial degradation under the same official-session PGD protocol. This supports an architecture-independent interpretation: bounded adversarial perturbations can produce task failure in neural-interface decoders even when the decoder is not a deep EEGNet. The classical results are not treated as Lipschitz-certified models; they are used to test whether the task-failure phenomenon generalizes beyond the certified compact neural model and beyond EEGNet.

The E1 label-permutation control checks that the original motor-imagery signal is not a trivial split or loader artifact. In this control, training labels are permuted while test labels remain true. Mean clean accuracy collapses to 0.2176 across the nine subjects, near the four-class chance level of 0.25. This control supports the interpretation that the real E1 runs are learning task-relevant structure, even though that structure remains fragile under adversarial stress.

\subsection{E2: Proxy-Fidelity Divergence}

The central E2 finding is that no single auxiliary objective simultaneously preserves all fidelity dimensions: repairing one metric damages another. The official-session EEGNet proxy-fidelity audit compares a classification-only encoder against a joint classification-plus-reconstruction encoder under the same latent bottleneck. The classification-only encoder is paired with a frozen reconstruction probe, so the probe measures how much signal information remains recoverable after task-only training. The joint encoder is trained with an explicit reconstruction objective, allowing the latent space to adapt to a fidelity requirement.

\begin{figure}[t]
\centering
\includegraphics[width=\columnwidth]{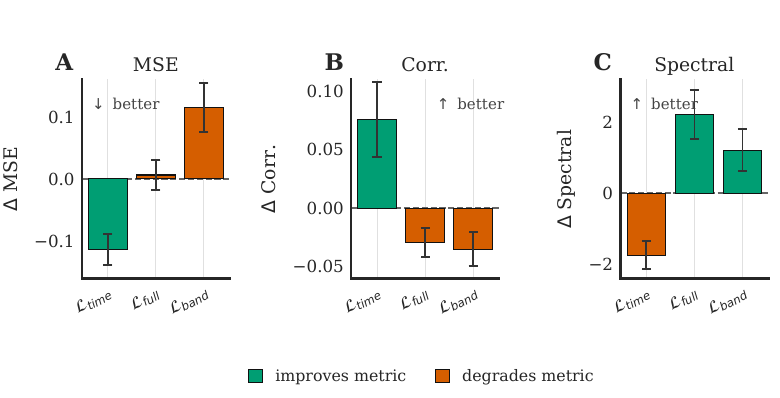}
\caption{Proxy-fidelity effects depend on the training objective. Time-domain losses improve MSE/correlation while worsening spectral log-MSE; spectral objectives reverse the spectral failure while trading away time-domain gains.}
\label{fig:e2_proxy_fidelity}
\end{figure}

The time-domain joint objective improves reconstruction MSE across all 9 official-session subjects and improves reconstruction correlation on average, while producing a small task-accuracy tradeoff (mean classification-only accuracy $0.4942$, joint accuracy $0.4776$; $\Delta = -0.0166$). Critically, the same objective worsens spectral log-MSE; the proxy-fidelity divergence in miniature. An auxiliary objective can repair one notion of fidelity while damaging another. Table~\ref{tab:e2_proxy_fidelity_objectives} summarizes the metric-specific deltas, and \Cref{fig:e2_proxy_fidelity} visualizes the MSE, correlation, and spectral log-MSE tradeoffs.

\begin{table}[t]
\centering
\caption{Proxy-fidelity deltas for three E2 objectives.}
\label{tab:e2_proxy_fidelity_objectives}
\footnotesize
\resizebox{\columnwidth}{!}{%
\begin{tabular}{lccc}
\toprule
Objective & $\Delta$ MSE & $\Delta$ corr. & $\Delta$ spectral log-MSE \\
\midrule
Time-MSE & $\mathbf{-0.1132 \pm 0.0252}$ & $\mathbf{0.0756 \pm 0.0321}$ & $\mathbf{1.7479 \pm 0.3994}$ \\
Full spectral & $0.0062 \pm 0.0234$ & $-0.0294 \pm 0.0126$ & $\mathbf{-2.2154 \pm 0.6945}$ \\
Bandpower & $0.1152 \pm 0.0395$ & $-0.0350 \pm 0.0143$ & $-1.2113 \pm 0.5973$ \\
\bottomrule
\end{tabular}%
}
\vspace{0.25em}
\parbox{\columnwidth}{\footnotesize \emph{Note.} Deltas are joint objective minus classifier-probe baseline; negative MSE/spectral and positive correlation indicate improvement.}
\end{table}

The spectral objectives provide the second half of the result. Adding a full-spectrum spectral loss reverses the earlier spectral failure, producing a large improvement in spectral log-MSE relative to the classifier-only frozen probe. The bandpower objective also improves spectral log-MSE, but less strongly under the full-spectrum evaluation metric. Both spectral objectives trade away the time-domain MSE/correlation gains produced by the time-domain-only objective. The spectral panel in \Cref{fig:e2_proxy_fidelity} visualizes this reversal.

The conclusion is objective-dependent: proxy-fidelity audits should ask \emph{which} fidelity metric is being preserved, which is being sacrificed, and whether the selected metric corresponds to the deployment objective. The E2 label-permutation control reinforces this interpretation; reconstruction improvements survive label randomization because reconstruction uses true neural epochs, not class semantics. E2 is a representation-fidelity diagnostic, not a classifier-performance comparison.

\subsection{E3: Latent Privacy Leakage}

The E3 experiment tests whether a public-task encoder leaks private subject identity. Under the SEED-IV session-holdout protocol, the baseline linear subject probe recovers identity with accuracy 0.4805, far above the 15-subject chance level of approximately 0.0667. Because train and test sessions are separated, this leakage cannot be explained solely by nearby windows from the same recording session appearing in both splits. It indicates that the learned latent representation carries subject-specific information that persists across sessions.

\begin{figure*}[t]
\centering
\includegraphics[width=0.92\textwidth]{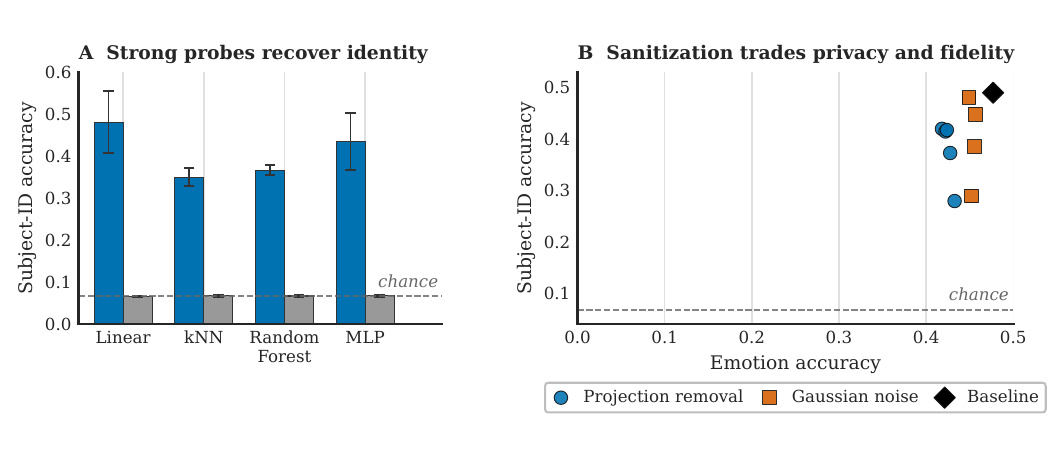}
\caption{Latent privacy leakage under the SEED-IV session-holdout protocol. Stronger probes recover subject identity far above chance, while the private-attribute permutation control collapses to chance; projection removal reduces leakage while preserving public-task accuracy.}
\label{fig:e3_privacy_leakage}
\end{figure*}

The stronger-probe suite tests whether this is merely a linear-probe artifact. kNN, random forest, and MLP subject probes also recover identity above chance, as summarized in Table~\ref{tab:e3_stronger_privacy_probes} and \Cref{fig:e3_privacy_leakage}. The nonlinear and nonparametric probes are deliberately bounded for tractability, but their above-chance performance strengthens the privacy claim: subject information is not visible only to one fragile linear classifier.

\begin{table}[t]
\centering
\caption{Stronger subject-identity probes and E3 null control.}
\label{tab:e3_stronger_privacy_probes}
\scriptsize
\resizebox{\columnwidth}{!}{%
\begin{tabular}{lcccccc}
\toprule
Probe & Real acc. & Perm. acc. & Real rem-12 & Perm. rem-12 & Real noise-1 & Perm. noise-1 \\
\midrule
Linear & $\mathbf{0.4805 \pm 0.0740}$ & $0.0663 \pm 0.0014$ & $\mathbf{0.1823 \pm 0.0969}$ & $-0.0005 \pm 0.0006$ & $\mathbf{0.1393 \pm 0.0457}$ & $0.0005 \pm 0.0018$ \\
kNN & $0.3495 \pm 0.0218$ & $0.0669 \pm 0.0023$ & $0.0855 \pm 0.0123$ & $-0.0003 \pm 0.0014$ & $0.0650 \pm 0.0069$ & $0.0033 \pm 0.0004$ \\
Random forest & $0.3657 \pm 0.0133$ & $0.0665 \pm 0.0014$ & $0.0261 \pm 0.0246$ & $-0.0005 \pm 0.0018$ & $0.0692 \pm 0.0225$ & $0.0000 \pm 0.0030$ \\
MLP & $0.4347 \pm 0.0674$ & $0.0671 \pm 0.0019$ & $0.1108 \pm 0.0657$ & $0.0020 \pm 0.0005$ & $0.1211 \pm 0.0462$ & $0.0007 \pm 0.0059$ \\
\bottomrule
\end{tabular}%
}
\vspace{0.25em}
\parbox{\columnwidth}{\footnotesize \emph{Note.} Values are mean $\pm$ SD over three seeds; chance is $1/15=0.0667$, and bold marks the strongest leakage/reduction.}
\end{table}

The sanitizer curves evaluate whether simple mechanisms can reduce leakage while preserving public-task utility. Projection removal deletes subject-discriminative latent directions, and Gaussian latent noise perturbs standardized embeddings. Removing 12 subject directions reduces leakage for the linear, kNN, random-forest, and MLP probes, with the strongest reduction under the linear probe. The aggregate emotion-accuracy delta for remove-12 is small and positive in the stronger-probe run, which makes projection removal the cleaner privacy-fidelity intervention in this experiment. \Cref{fig:e3_privacy_leakage} shows the leakage reduction pattern.

The private-attribute permutation control is the key negative control. When subject labels are permuted within the real session-holdout split, baseline subject-probe accuracy collapses to chance for all four probe families: the permuted baselines are approximately 0.066-0.067. Leakage reductions also become near zero. This rules out the simplest artifact explanations, including probe capacity, class imbalance, or the mere existence of the session-holdout split. The E3 result is therefore best read as evidence that public neural embeddings can retain unintended private information, and that empirical privacy probes are necessary even when the public task appears well defined.

\subsection{Physiological Robustness}

The physiological robustness suite extends E1 beyond optimized PGD attacks by evaluating official-session EEGNet models under perturbations that resemble deployment-time failures: channel dropout, Gaussian sensor noise, band-limited noise, temporal shift, and amplitude scaling. These are applied after preprocessing and per-epoch standardization, so they are model-input stress tests rather than raw-device simulations.

Strong Gaussian sensor noise, channel dropout, and band-specific noise produce accuracy drops and prediction-flip fractions comparable to the PGD results across all nine subjects. High-beta and low-gamma band-limited perturbations are especially damaging, indicating that frequency-local disruptions can be consequential even without adversarial optimization. Temporal shifts produce smaller accuracy effects but still increase prediction instability.

\begin{table}[t]
\centering
\caption{Largest physiological robustness failures for official-session EEGNet.}
\label{tab:physiological_robustness}
\scriptsize
\resizebox{\columnwidth}{!}{%
\begin{tabular}{llcccc}
\toprule
Perturbation & Severity & Pert. acc. & Acc. drop & Flip frac. & $n$ \\
\midrule
High-beta noise & 0.50 & $0.2755 \pm 0.0392$ & $\mathbf{0.3187 \pm 0.1718}$ & $\mathbf{0.7172 \pm 0.0839}$ & 9 \\
Gaussian noise & 0.50 & $0.2851 \pm 0.0402$ & $\mathbf{0.3090 \pm 0.1536}$ & $0.6925 \pm 0.0836$ & 9 \\
Low-gamma noise & 0.50 & $0.2932 \pm 0.0724$ & $\mathbf{0.3009 \pm 0.1574}$ & $0.6863 \pm 0.1168$ & 9 \\
Amplitude scale & 0.50 & $0.2982 \pm 0.0401$ & $\mathbf{0.2959 \pm 0.1448}$ & $0.6416 \pm 0.1620$ & 9 \\
Channel dropout & 0.50 & $0.3067 \pm 0.0407$ & $\mathbf{0.2874 \pm 0.1398}$ & $0.6474 \pm 0.0510$ & 9 \\
Alpha noise & 0.50 & $0.3117 \pm 0.0706$ & $\mathbf{0.2824 \pm 0.1937}$ & $0.6227 \pm 0.1700$ & 9 \\
High-beta noise & 0.25 & $0.3391 \pm 0.0876$ & $0.2550 \pm 0.1610$ & $0.5903 \pm 0.1522$ & 9 \\
Channel dropout & 0.25 & $0.3495 \pm 0.0512$ & $0.2446 \pm 0.1312$ & $0.6038 \pm 0.0534$ & 9 \\
Low-gamma noise & 0.25 & $0.4317 \pm 0.1522$ & $0.1624 \pm 0.1262$ & $0.4518 \pm 0.2063$ & 9 \\
Gaussian noise & 0.25 & $0.4325 \pm 0.1428$ & $0.1617 \pm 0.1147$ & $0.4668 \pm 0.1584$ & 9 \\
Channel dropout & 0.10 & $0.4383 \pm 0.0995$ & $0.1559 \pm 0.0781$ & $0.4217 \pm 0.0590$ & 9 \\
Alpha noise & 0.25 & $0.4564 \pm 0.1093$ & $0.1377 \pm 0.1282$ & $0.3862 \pm 0.1545$ & 9 \\
Amplitude scale & 1.50 & $0.4715 \pm 0.1393$ & $0.1227 \pm 0.1013$ & $0.4047 \pm 0.1536$ & 9 \\
Amplitude scale & 0.75 & $0.4811 \pm 0.1592$ & $0.1130 \pm 0.0834$ & $0.3661 \pm 0.1032$ & 9 \\
Amplitude scale & 1.25 & $0.5370 \pm 0.1453$ & $0.0571 \pm 0.0639$ & $0.2681 \pm 0.1218$ & 9 \\
Gaussian noise & 0.10 & $0.5675 \pm 0.1617$ & $0.0266 \pm 0.0329$ & $0.1617 \pm 0.0838$ & 9 \\
Temporal shift & 32 & $0.5768 \pm 0.1646$ & $0.0174 \pm 0.0201$ & $0.1246 \pm 0.0714$ & 9 \\
Temporal shift & 8 & $0.5883 \pm 0.1664$ & $0.0058 \pm 0.0148$ & $0.0451 \pm 0.0354$ & 9 \\
\bottomrule
\end{tabular}%
}
\vspace{0.25em}
\parbox{\columnwidth}{\footnotesize \emph{Note.} Values are mean $\pm$ SD across subjects; bold marks physiological drops at or above the EEGNet PGD $\varepsilon=0.25$ drop of $0.2573$.}
\end{table}

\begin{figure}[t]
\centering
\includegraphics[width=\columnwidth]{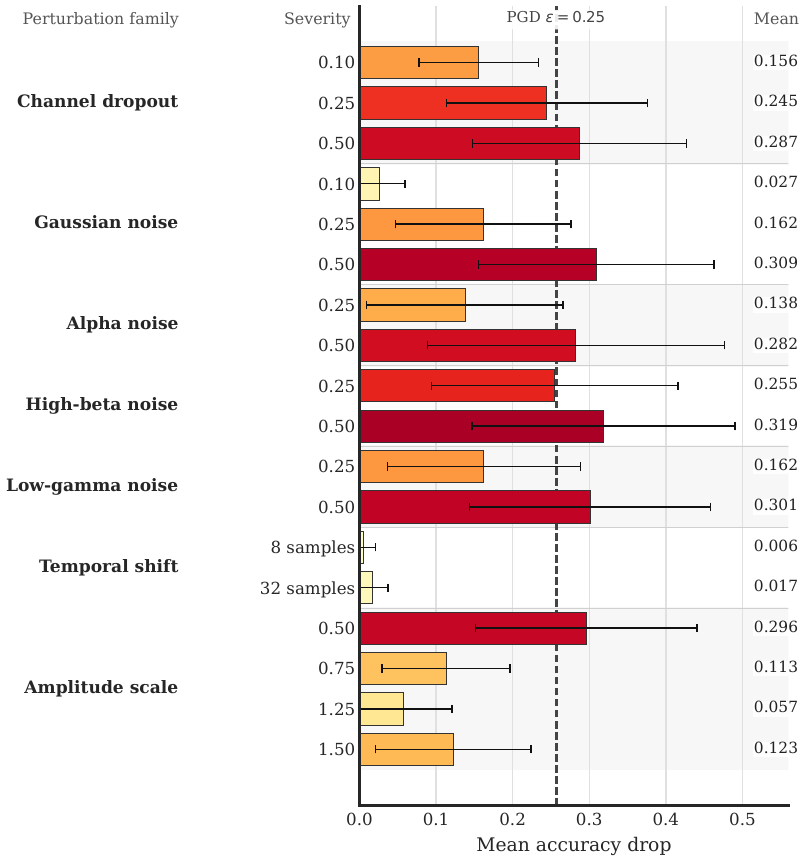}
\caption{Physiological robustness stress tests. Several non-PGD perturbations produce mean accuracy drops comparable to the official-session EEGNet PGD drop at $\epsilon=0.25$ ($0.2573$).}
\label{fig:physiological_robustness_top_drops}
\end{figure}

Table~\ref{tab:physiological_robustness} and Figure~\ref{fig:physiological_robustness_top_drops} position these stress tests as operational evidence rather than as a replacement for PGD. PGD shows that worst-case bounded perturbations can collapse task performance; the physiological suite shows that plausible non-adversarial sensor and signal changes can also degrade behavior. Together, they ground the safety argument in the conditions under which embedded neural interfaces are likely to fail: noisy channels, unstable amplitudes, frequency-local artifacts, and brittle decoder responses.

\subsection{Summary of Statistical Evidence}

Table~\ref{tab:statistical_summary} consolidates the inferential results across all three audits.

\begin{table}[t]
\centering
\caption{Statistical summary across audits. All E1/E2 tests use $n{=}9$ subject-paired sign-flip permutations; E3 uses $n{=}3$ seeds. CI: bootstrap 95\%; $q$: BH-corrected $p$-value.}
\label{tab:statistical_summary}
\footnotesize
\begin{tabular}{@{}llll@{}}
\toprule
Audit & Effect & Mean [CI] & $p$ / $q$ \\
\midrule
E1 EEGNet & Acc.\ drop, $\epsilon{=}0.25$ & $.257$ $[.224,.296]$ & $.002/.002$ \\
E1 CSP+LDA & Acc.\ drop, $\epsilon{=}0.25$ & $.335$ $[.302,.368]$ & $.002/.002$ \\
E1 FBCSP & Acc.\ drop, $\epsilon{=}0.25$ & $.244$ $[.212,.274]$ & $.002/.002$ \\
\midrule
E2 Time & MSE $\Delta$ & $-.113$ $[-.128,-.097]$ & $.002/.004$ \\
E2 Time & Spec.\ log-MSE $\Delta$ & $+1.748$ $[1.50,1.99]$ & $.002/.004$ \\
E2 Spectral & Spec.\ log-MSE $\Delta$ & $-2.215$ $[-2.67,-1.82]$ & $.002/.004$ \\
\midrule
E3 Linear & Subj.\ acc.\ (baseline) & $.481$ vs.\ $.067$ & -- \\
E3 Remove-12 & Leakage reduction & $.182$ $[.072,.253]$ & $.125$ \\
\bottomrule
\end{tabular}
\end{table}

The E1 accuracy drops are the strongest inferential claims in the paper: all three decoder families show large, statistically significant degradation under bounded PGD perturbation while the output-sensitivity certificate passes. The effect is architecture-independent and robust to the official train-to-evaluation session split across all nine BCI IV 2a subjects.

E2 confirms that proxy fidelity is not a scalar property. The time-domain joint objective improves MSE while worsening spectral log-MSE; the spectral objective reverses this tradeoff. The audit exposes \emph{which} fidelity notion each objective preserves, not whether fidelity is preserved in the abstract.

E3 provides consistent directional evidence for latent privacy leakage across all four probe families, with the private-attribute permutation control collapsing to chance. Because E3 uses only three seeds, the minimum exact one-sided $p$-value is $0.125$; we treat E3 as multiseed directional evidence rather than a conventionally significant hypothesis test.

\section{Discussion}\label{sec:discussion}

\subsection{Implications for Neural Interface Governance}

The results point to a governance lesson that is narrower than product regulation but stronger than ordinary benchmarking: neural-interface deployment should require empirical safety audits that test whether formal, proxy, and privacy claims survive operational stress.

The \emph{verification} dimension is directly implicated by E1. A mathematically valid certificate can fail to certify the task property users care about; certificates should therefore be paired with adversarial task-failure audits, margin checks, physiological perturbation tests, and null controls. The \emph{frontier-safety} dimension is implicated by E2: as neural-interface models scale toward richer adaptive representations, governance should ask not only whether a model is accurate, but which fidelity target it preserves and under which evaluation protocol. The \emph{agent-governance} dimension is implicated by E3, because latent states may contain private attributes that are not visible in the intended output and that persist across recording sessions.

This is a methodological claim, not a product proposal: neural-interface safety evidence should include a formal or mechanistic check, an operational stress test, a fidelity audit, a privacy-leakage audit, and negative controls that rule out artifact explanations.

\subsection{Limitations}

Several constraints bound the conclusions that can be drawn from the current evidence.

The experimental scope covers two noninvasive EEG datasets: BCI Competition IV 2a (motor imagery, 9 subjects) and SEED-IV (emotion recognition, 15 subjects). Both are well-established benchmarks, but neither captures the complexity of invasive neural interfaces, high-density recording arrays, or online adaptive decoding scenarios. The SEED-IV privacy experiments further use the pre-extracted \texttt{de\_LDS} feature representation rather than raw EEG, so E3 measures leakage in a feature space that is already substantially compressed relative to the original signal.

The Lipschitz certificate used in E1 is a global product-of-layer upper bound, which is known to be conservative. The resulting verification gap is therefore partly a consequence of bound looseness rather than a fundamental impossibility of certification. Tighter local Lipschitz estimation methods may narrow this gap, though whether they close it entirely for EEG decoders remains untested.

Statistical power is uneven across experiments. E1 and E2 use $n{=}9$ subject-paired tests and achieve conventionally significant $p$-values. E3 uses only three seeds, so the smallest achievable one-sided $p$-value is $0.125$; E3 claims are therefore directional evidence rather than formally significant hypothesis tests. The RF and MLP probes in E3 are also capacity-bounded and class-balanced subsampled, so the measured leakage may underestimate what a fully optimized adversarial probe could extract.

The physiological robustness suite applies perturbations after preprocessing and per-epoch standardization, making them model-input stress tests rather than end-to-end device simulations. Real deployment noise enters before preprocessing and interacts with filtering, artifact rejection, and re-referencing in ways that post-hoc perturbations do not capture.

Finally, the sanitization mechanisms evaluated in E3 (projection removal and Gaussian latent noise) are empirical interventions without formal privacy guarantees. They reduce probe accuracy but do not provide differential privacy accounting or worst-case leakage bounds.

\subsection{Open Problems}

First, the Lipschitz evidence is deliberately conservative. The E1 certificate is valid as an output-movement upper bound, but it is too loose to provide a useful nonzero margin certificate in the official-session PGD audit. A concrete direction is to replace global product-of-layer bounds with tighter local Lipschitz estimates around the data manifold, then test whether the resulting constants produce nonvacuous certified-safe fractions under the same PGD and physiological perturbation grids.

Second, the E3 sanitizers are empirical rather than formal privacy mechanisms. Projection removal and Gaussian latent noise reduce subject-probe accuracy, but they do not provide differential privacy accounting. A next step is to define the latent release mechanism explicitly, estimate or bound its privacy parameters, and compare those guarantees against the same linear, kNN, random-forest, and MLP probes. This matters because the current stronger-probe experiment uses only three seeds, bounded RF/MLP capacities, and class-balanced probe subsampling for tractability.

Third, E2 exposes a multi-objective tradeoff that remains unresolved. Time-domain reconstruction improves MSE and correlation while worsening spectral log-MSE; spectral objectives improve spectral log-MSE while trading away time-domain gains. The natural follow-up is to estimate a three-way Pareto surface over classification accuracy, time-domain fidelity, and spectral fidelity, using weight sweeps across multiple seeds and official-session subjects.

Finally, the evidence base should be extended to larger and more modern neural-interface settings. The current BCI evidence uses all nine BCI IV 2a subjects and the official protocol, but it remains limited to noninvasive EEG motor imagery and SEED-IV emotion features. Future audits should test larger BCI datasets, additional Braindecode architectures, raw-signal privacy settings, adaptive decoders, and online calibration scenarios.

\section{Conclusion}\label{sec:conclusion}

We introduced a unified empirical audit framework for embedded neural-interface models, grounded in three alignment-inspired failure modes: verification insufficiency, proxy-fidelity divergence, and latent information exfiltration. The framework connects robustness certification, signal fidelity, and privacy leakage as related consequences of optimizing an incomplete training objective rather than as isolated benchmark problems.

The central result is that conservative safety certificates alone are insufficient for neural-interface deployment. A Lipschitz-style output-sensitivity check passes while classification accuracy collapses by 25.7\% under bounded adversarial perturbation; an architecture-independent finding that holds across EEGNet, CSP+LDA, and FBCSP+LDA on the official BCI IV 2a evaluation protocol. E2 and E3 show analogous failures: proxy objectives preserve some signal properties while damaging others, and public-task embeddings retain private subject identity at $7.2\times$ the chance level. Neural-interface safety claims should not rest on certificates, clean accuracy, or intended-task performance alone. Operational auditing across verification, fidelity, and privacy is necessary.

\bibliographystyle{unsrt}
\bibliography{references}

\end{document}